\documentclass[10pt,twocolumn,letterpaper]{article}

\usepackage[pagenumbers]{cvpr} %

\usepackage[T1]{fontenc} %

\usepackage{tikz}
\usepackage{marvosym}
\usetikzlibrary{positioning,arrows.meta,calc}

\DeclareMathOperator*{\pose}{\boldsymbol{\theta}}

\definecolor{cvprblue}{rgb}{0.21,0.49,0.74}
\usepackage[pagebackref,breaklinks,colorlinks,allcolors=cvprblue]{hyperref}
\hypersetup{pdftitle={HARI4D: Hand-Articulation Aware 4D Reconstruction of Human-Object Interaction},
  pdfauthor={Semin Kim, Xianghui Xie, Chuqiao Li, Jongyoo Kim, Gerard Pons-Moll}}
\ifdefined\pdfsuppressptexinfo\pdfsuppressptexinfo=-1\fi

\title{HARI4D: \underline{H}and-\underline{A}rticulation Aware \underline{4D} \underline{R}econstruction of \\Human-Object \underline{I}nteraction}

\author{
Semin Kim$^{1,*}$\quad
Xianghui Xie$^{2,3,4}$\textsuperscript{\Letter}\quad
Chuqiao Li$^{2,3}$\quad
Jongyoo Kim$^{1}$\quad
Gerard Pons-Moll$^{2,3,4}$\\[0.5em]
{\normalsize
$^{1}$Yonsei University, South Korea\quad
$^{2}$University of T\"{u}bingen, Germany\quad
$^{3}$T\"{u}bingen AI Center, Germany}\\
{\normalsize
$^{4}$Max Planck Institute for Informatics, Saarland Informatics Campus, Germany}\\[0.3em]
{\small $^{*}$Work partially done during a visit to the University of T\"{u}bingen.\quad
\Letter~Corresponding author.}\\
{\small\href{https://virtualhumans.mpi-inf.mpg.de/hari4d/}{https://virtualhumans.mpi-inf.mpg.de/hari4d/}}
}

\begin{document}
\newcommand{\XH}[1]{{\color{blue}[XH: #1]}}
\newcommand{\reachnet}[1]{{ReachNet}}
\newcommand{\touchnet}[1]{{TouchNet}}
\newcommand{\method}[1]{{HARI4D}}
\makeatletter
\twocolumn[{%
  \@maketitle
  \vspace{-22pt}
  \begin{center}
    \includegraphics[width=\textwidth]{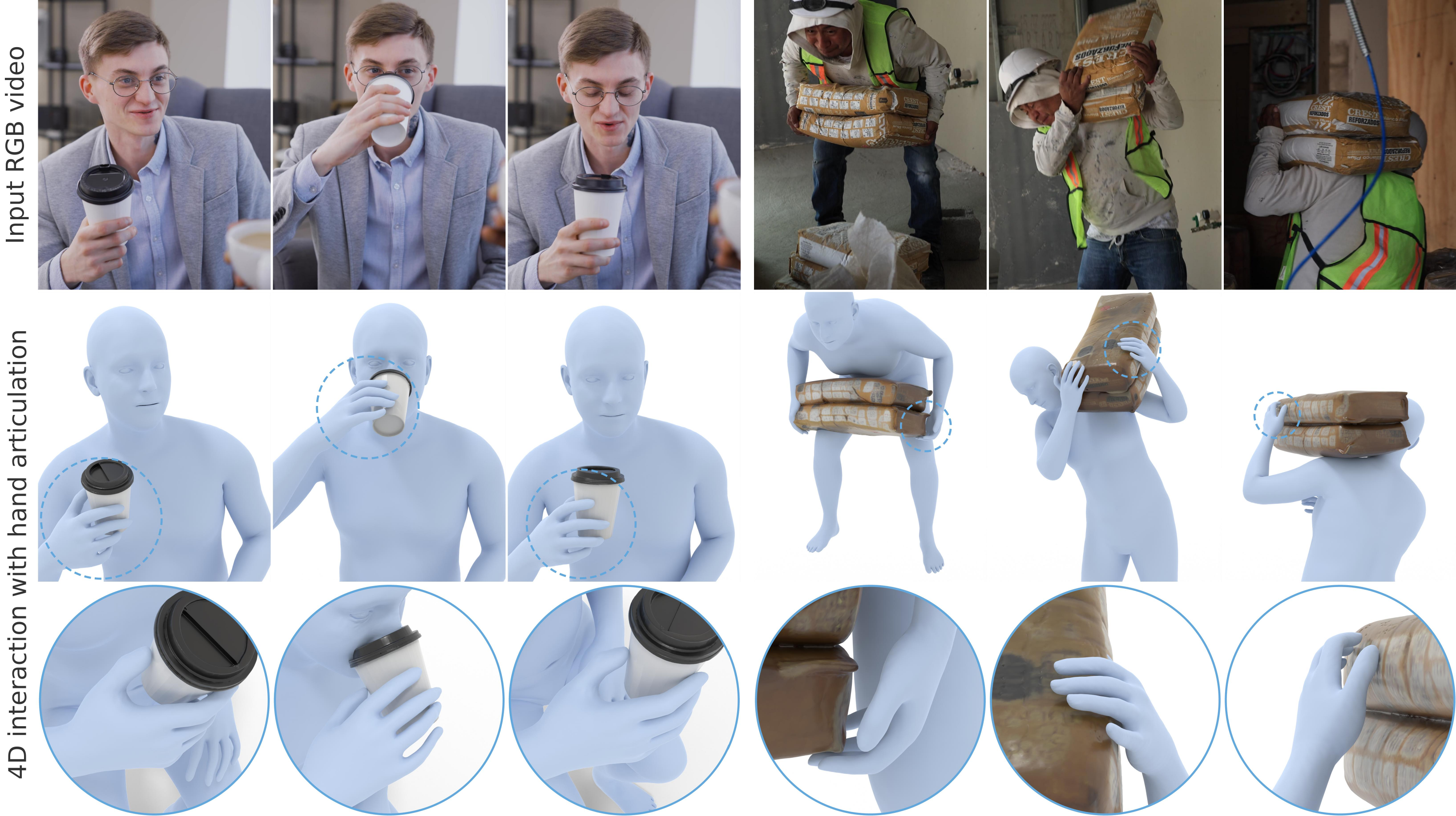}
    
    \captionof{figure}{Given a monocular RGB video, \method{} reconstructs the 4D human-object interaction with hands. Our method recovers the holistic interaction from full body down to articulated hands.
      }
    \label{fig:teaser}
  \end{center}
  \vspace{4pt}
}]
\makeatother

\begin{abstract}
Reasoning about human-object interaction from RGB cameras is important for gaming, VR environments, and robot learning. Yet 4D reconstruction from a monocular RGB camera is challenging, especially when articulated hands are involved. Prior full-body works neglect hands during reconstruction, causing them to penetrate into, or float from the object. We present HARI4D, a hand-articulation-aware method for reconstructing spatially and temporally consistent 4D human--object interaction from monocular RGB video. We introduce a history-aware object tracker that provides robust and stable tracking. Based on these tracks, we improve the interaction configuration through coarse-to-fine reasoning that follows the order of object manipulation: \reachnet{} first brings the full body and arms to the object during contact, and \touchnet{} then finalizes hand articulation according to temporal context and local object geometry. On ARCTIC, unseen in training, our method reduces the object error by 56\% and improves hand contact F1 from 0.866 to 0.921 over prior art. Our method recovers accurate full body-object interaction with coherent hand articulations and applies to online videos. 
Our code and pretrained models will be released.
\end{abstract}
    
\section{Introduction}
\label{sec:intro}

Reconstructing human--object interactions from ordinary RGB videos can turn demonstrations into rich supervision for embodied agents. Beyond identifying which object moves, a robot must understand how the body approaches it, the arm positions the wrist, and the fingers establish and maintain a grasp. Metric 4D reconstruction of this process can support robot learning from human demonstrations~\cite{zhao2026real2sim,huang2026physicalhoi,xie2026grail}, realistic digital humans for games~\cite{grassal2022neuralhead}, and interactive virtual and augmented reality content~\cite{orts2016holoportation}.

Recent full-body human--object interaction (HOI) methods have progressively relaxed their assumptions on the manipulated object: from template-based reconstruction~\cite{xie2023vistracker, zhang2020perceiving}, to reconstructing unseen instances within specific categories~\cite{xie2025intertrack, cseke2025pico}, and finally to category-agnostic reconstruction~\cite{xie2026cari4d,wen2025reconstructing}. Yet these methods represent the hands without detailed hand articulation, precisely where physical manipulation happens. Conversely, hand--object methods recover articulated hands from images~\cite{hasson2019obman,ye2022ihoi,chen2023gsdf,liu2025easyhoi} or videos~\cite{hasson2021homan,ye2023diffhoi,fan2024hold,wang2025magichoi}, but typically leave out the supporting full-body pose. This separation is limiting: a plausible object manipulation not only depends on hand and wrist poses but also should be supported by the arm and full body. 

Bridging these two scales is particularly difficult in monocular video. Depth ambiguity makes the relative placement of the human and object underconstrained, while their motion causes persistent and changing occlusion. The evidence also becomes progressively weaker toward the point of contact: the body occupies a large image region, but the wrist and fingers may span only a few pixels and are often hidden by the object during manipulation. A successful method must therefore reason jointly about body motion, object motion, and hand articulation, while using temporal context to recover what is invisible in individual frames.

We introduce \textbf{\underline{H}}and-\textbf{\underline{A}}rticulation Aware \textbf{\underline{4D}} \textbf{\underline{R}}econstruction of Human--Object \textbf{\underline{I}}nteraction (\method{}), a category-agnostic method for reconstructing metric 4D full-body HOI with articulated hands from a monocular RGB video. Our key idea is to follow the natural hierarchy of manipulation: the body first approaches the object, the arm then positions the wrist, and the fingers finally form the grasp. We initialize the human pose and metric object shape with complementary foundation models, and obtain a stable object anchor with our history-conditioned object pose tracker. We then refine the interaction in a coarse to fine manner. \reachnet{} uses local object geometry, image-based contact cues, and temporal context to adjust the body, arms, and wrists. Given this holistic reconstruction, \touchnet{} reasons over the local object surface through time to refine wrist orientation and hand articulation. Because both refiners operate on local geometric relations rather than object-specific features, the resulting system generalizes to unseen objects, see \cref{fig:teaser}.

We evaluate \method{} on ARCTIC~\cite{fan2023arctic} and BEHAVE~\cite{bhatnagar22behave}. Compared with prior art CARI4D, our method reduces combined Chamfer distance by 42.9\%, while improving hand-contact F1 from 0.866 to 0.921 on ARCTIC. On BEHAVE, it reduces combined Chamfer distance by 11.8\%, human acceleration error by 57.8\%, and object acceleration error by 54.0\%. Our ablations further show that our tracker substantially improves object tracking, while \reachnet{} brings better body-object arrangement and \touchnet{} improves local hand-object interaction geometry. 

In summary, our main contributions are:
\begin{itemize}
    \item We present \method{}, a category-agnostic method to reconstruct metric 4D full-body human--object interactions with articulated hands from monocular RGB video.
    \item We introduce a history-conditioned object pose refiner that incorporates past motion into current pose reasoning, yielding more accurate and temporally stable object tracking under occlusion.
    \item We propose a hierarchical interaction-reasoning framework comprising \reachnet{} for contact-aware body, arm, and wrist refinement and \touchnet{} for temporally consistent hand articulation refinement. Together with our tracker, they outperform prior work in reconstruction error on both ARCTIC and BEHAVE.
\end{itemize}
Our code and pretrained models will be released.

\section{Related Work}
\label{sec:related_works}

\textbf{3D foundation models.} Human mesh recovery now handles in-the-wild images~\cite{yang2026sam, patel2025camerahmr, sarandi2024neural}. Image-to-3D models generate object shape from a single image~\cite{chen2026sam, zhao2025hunyuan3d, xiang2026native}, and pose estimators recover object pose from RGB or RGB-D input~\cite{wen2024foundationpose, lee2025any6d}. Single-image geometry models predict relative depth~\cite{wang2025moge, depthanything} or metric depth~\cite{wang2025moge2, piccinelli2024unidepth, piccinelli2025unidepthv2, kong2026finedetailmonoculargeometryestimation}, and reconstruction models recover whole scenes from multiple views~\cite{wang2025vggt, wang2026vggt, depthanything3}. While these methods show impressive performance in their respective goals, they do not reason about the relationship and interaction between components.

\noindent\textbf{Hand-object interaction and grasp modeling.} Hands and hand-held objects have been reconstructed from single images~\cite{hasson2019obman, ye2022ihoi, chen2023gsdf, liu2025easyhoi} and from videos~\cite{hasson2021homan, ye2023diffhoi, fan2024hold, wang2025magichoi}, often down to finger articulation, but these methods leave out the body. Other works focus on the hand-object interaction itself using implicit fields, contact models and generative priors~\cite{karunratanakul2020graspingfield, liu2023contactgen, ye2024ghop} that describe how a hand fits an object, ContactOpt~\cite{grady2021contactopt}, TOCH~\cite{zhou2022toch} and GeneOH Diffusion~\cite{liu2024geneoh} refine noisy hand poses or motion against the object, and GEARS~\cite{zhou2024gears} reasons about the local object geometry near each hand joint. These methods are successful in finding accurate finger articulations, but do not consider the full body motion which is critical to finding a plausible wrist position.

\noindent\textbf{Full body human and interaction.} Human-scene interaction reconstruction models the person with the surrounding geometry~\cite{hassan2019prox, yi2022mover} or by recovering people and scenes together~\cite{muller2025hsfm, liu2025josh, chen2026human3r}. PhySIC~\cite{muralidhar2025physic} optimizes a body against a single reconstructed scene with contact and interpenetration terms, and GRAFT~\cite{ym2026graft} uses a learned iterative refiner that probes the scene geometry around the body.

Single-image human-object interaction reconstruction \cite{xie2024rhobin} has been studied through template-based fitting~\cite{zhang2020perceiving, xie2022chore, nam2024joint, li2025scorehoi, wang2025end}, template-free reconstruction~\cite{xie2024template, wen2025reconstructing, chatterjee2026milo, chen2026reconstructing}, mesh retrieval~\cite{cseke2025pico, dwivedi2025interactvlm}, and learned interaction priors~\cite{antic2026lexis}. On video, VisTracker~\cite{xie2023vistracker} tracks with a known template, InterTrack~\cite{xie2025intertrack} removes the template while still being category bound. CARI4D~\cite{xie2026cari4d} generalizes to unseen categories by building on foundation models, but it does not regress detailed finger articulation. Other works add an IMU on the object~\cite{zhao2024imhoi}, a physics simulator~\cite{huang2026physicalhoi}, a moving camera~\cite{xue2026rhino}, lift 2D keypoints to world-space body and object motion~\cite{li2026anylift}, or sparse head and wrist tracking~\cite{petrov2026echo}. HA-HOI~\cite{zhao2026real2sim} fuses hand estimates into the body and refines the wrists and fingers against the object by optimization, and GraspHOI~\cite{kim2026grasphoi} likewise optimizes the arms and fingers against the object surface, but from a single image.
Full-body motion \cite{li2025unimotion,li2026frankenmotion,nazarenus2026actionplan} and grasping \cite{berndt2026pihand} have also been studied generatively~\cite{tendulkar2023flex, zheng2023coop, paschalidis2025wholebody, taheri2022goal, wu2022saga, li2026pami, sharma_and_xie2026Hoi3DGen}, synthesizing whole-body grasping motion, and GRIP~\cite{taheri2024grip} denoises the arms and synthesizes the fingers from a given body and object motion. Unlike HA-HOI and GraspHOI, which do heavy test time optimization, \method{} refines the body, arms and hands with an efficient feedforward network. 

\begin{figure*}[ht]
    \centering
    \includegraphics[width=1.0\linewidth]{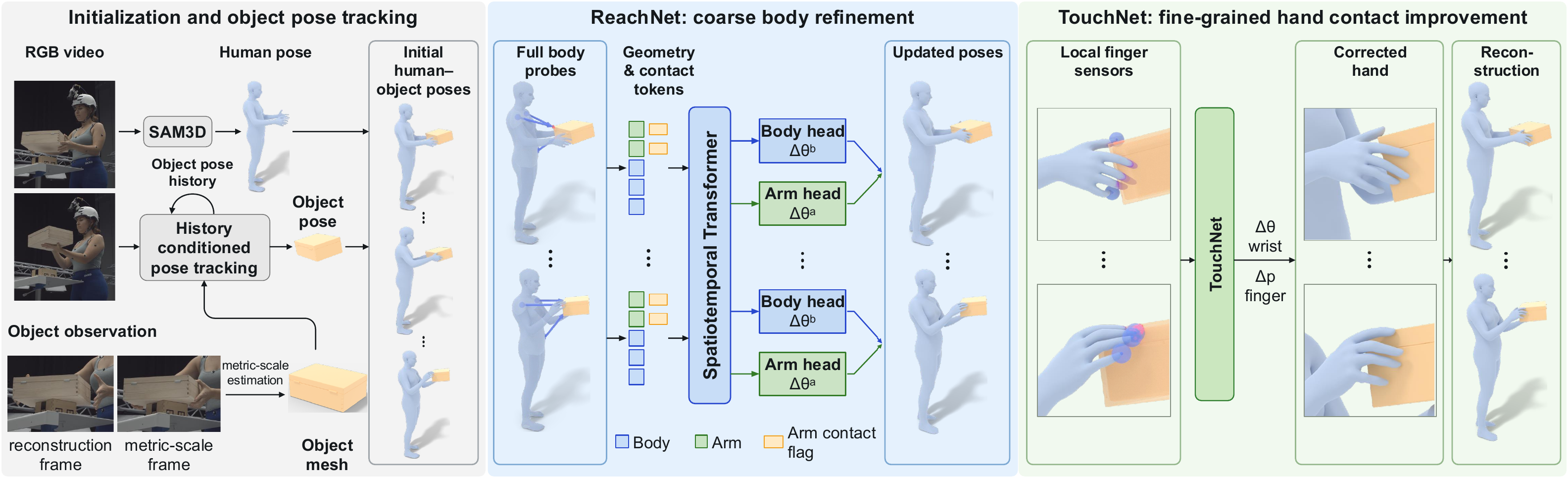}
    \caption{\textbf{Method Overview.} Starting from foundation models, we initialize the human pose and metric-scale object mesh (\cref{sec:method_init}). We then train a history conditioned object pose tracker to robustly track the object pose under dynamic movement occlusions (\cref{sec:method_track}). We obtain plausible full body and hand articulations via coarse to fine interaction reasoning (\cref{sec:method_refinement}), starting from \reachnet{} that brings the human and arms to the object when necessary, and then \touchnet{} which refines hand-object surface contacts.
    }
    \label{fig:method}
\end{figure*}
\providecommand{\chg}[1]{\leavevmode{\color{chgcolor}#1}}
\providecolor{chgcolor}{RGB}{72,82,168}
\section{Method}
\label{sec:method}
Given an RGB video $\{I_t\}_{t=1}^{T}$ recorded by a monocular camera, our goal is to recover the metric 4D motion of a person and a rigid object, including the person's articulated hands. We represent the person by SMPL-X~\cite{pavlakos2019expressive} and the object by a mesh $\mathcal{O}$ with a 6DoF pose $({R}_t,\boldsymbol{\tau}_t)$ in every frame. In addition to the video, we take binary person and object masks, $M^h_t$ and $M^o_t$ respectively, as input.

This setting requires reasoning at two very different spatial scales. Full-body motion is supported by large image regions, whereas the wrist and fingers occupy few pixels and are frequently occluded when contact occurs. We therefore reconstruct the interaction hierarchically, mirroring the structure of manipulation: coarse body motion first brings a hand near the object, then the wrist and fingers establish the plausible grasp. We initialize the human and metric object in a common camera frame (\cref{sec:method_init}), obtain a stable object anchor with history-conditioned pose tracking (\cref{sec:method_track}), and finally refine the human against this fixed object from the body to the arms and articulated hands (\cref{sec:method_refinement}).

\subsection{Human Pose and Object Shape Initialization}
\label{sec:method_init}
We start with estimating per-frame human pose $\theta_t$ and object shape $\mathcal{O}$ whose scale is aligned with human-scale using the depth from MoGe-3~\cite{kong2026finedetailmonoculargeometryestimation}. Specifically, we run MoGe-3 to  estimate a metric depth map $\mathbf{D}_t$ and the camera intrinsics $K$, which are estimated on the first frame and then held fixed. Using the same $K$, SAM~3D~Body~\cite{yang2026sam} recovers a Momentum Human Rig~\cite{MHR:2025} in every frame from the bounding box of $M^h_t$. We convert it to SMPL-X with the official fitting procedure, obtaining the camera-frame translation, global orientation, body and hand poses.

The body and depth estimates nevertheless disagree by a time-varying depth offset. We align depth to the more stable body estimate using visible body regions that are considered reliable. 
HOI-DETR~\cite{darkhalil2026improving} provides a binary contact flag $c^s_t\in\{0,1\}$ for each hand $s\in\{\mathrm{L},\mathrm{R}\}$. For each frame, we denote $\mathcal{S}_t$ as the visible vertices of the body region for alignment, which is the hands when they are reliable and in contact with the object or the body otherwise. We align the MoGe-3 depth map to body depth by computing an offset $\delta_t$ from the difference between body vertex depth $\mathbf{v}_z$ and estimated depth: $\delta_t=\operatorname{median}_{{\mathbf{v}}\in\mathcal{S}_t}\Big(\mathbf{D}_t\big(\pi({\mathbf{v}})\big)-\mathbf{v}_z\Big)$, where $\pi$ denotes projection by $K$. Aligned depth is obtained by subtracting temporally smoothed $\delta_t$: $ \tilde{\mathbf{D}}_t=\mathbf{D}_t-\bar\delta_t$.

We further reconstruct 3D object shape and its metric-scale that aligns with the human scale. Observing that foundation depth estimators tend to produce good relative geometry when hand and object are in contact, we hence use a contacting frame to estimate the metric-scale and pick another nearby frame to reconstruct the object. Specifically, we first find an interval of still frames where object does not move. Then, searching from either end of this interval towards its middle, we use HOI-DETR~\cite{darkhalil2026improving} to find frame $A$, where the object is in contact with any hand and the hand depth overlaps with the object depth. The frame in this static interval with the largest object region visible is the rest frame $B$. We use frame $B$ to reconstruct the canonical mesh using SAM~3D~\cite{chen2026sam}, and estimate the metric-scale of the object with depth estimation from frame $A$ using FoundationPose~\cite{wen2024foundationpose}. For convenience, we denote the metric scaled object mesh as $\mathcal{O}$. More details on frame selection and scale estimation are provided in \cref{sec:supp_initialization}.

\subsection{History-conditioned Object Tracking}
\label{sec:method_track}
Our reconstruction uses the object as an anchor for the subsequent human refinement. We add history conditioning with FiLM~\cite{perez2018film} to the FoundationPose~\cite{wen2024foundationpose} refiner and run the tracking with the fine-tuned model and a velocity-based Kalman filter. This combines current image evidence with tracked history motion, producing more stable and robust object tracking than vanilla FoundationPose. 

FoundationPose refiner uses an image encoder to extract image feature which is then passed to rotation and translation prediction heads each consisting of one transformer encoder followed by one linear layer. This refiner sees only the current frame and the pose it starts from, not the motion that led there, so under occlusion or noisy depth it can jitter or produce a temporally inconsistent orientation.

\noindent\textbf{Motion-history features.}
We describe the motion before frame $t$ by the relative pose between nearby frames and concatenate them over up to $H=90$ preceding frames. Specifically we define the relative pose feature at a history frame $t-k$ as: $\boldsymbol{\delta}_{t-k}=\mathrm{clip}_{\pm5}\Big[\tfrac{\log\left({R}_{t-k}{R}_{t-k-1}^{\top}\right)}{\rho},\tfrac{\boldsymbol{\tau}_{t-k}-\boldsymbol{\tau}_{t-k-1}}{r}\Big]$, where ${R}_s$ and $\boldsymbol{\tau}_s$ are the object rotation and translation at frame $s$, and $\log(\cdot)$ returns the axis-angle vector of a rotation matrix. And $\rho,r$ are the normalization parameters defined in \cref{sec:supp_tracking}. The motion history feature $\mathbf{h}_t$ concatenates all relative pose features $\boldsymbol{\delta}_{s}$ after binary masking $\nu_s\in \{0, 1\}$: $\mathbf{h}_t=[\nu_{t-1}\cdot\boldsymbol{\delta}_{t-1},...,\nu_{t-H}\cdot\boldsymbol{\delta}_{t-H}]$, where $\nu_{t-k}$ is 1 when the poses at $t-k$ and $t-k-1$ are both available and 0 otherwise. 

We then modulate the feature token $\mathbf{h}_t$ into the refiner using FiLM~\cite{perez2018film}. Specifically, we do per-channel scale and shift of the output feature $\mathbf{F}_t$ from the transformer layer of FoundationPose refiner's decoder. We apply modulation to both rotation and translation decoders of the refiner. This history conditioning improves the robustness of object pose tracking, see~\cref{tab:ablation}. 

We fine-tune all weights on renderings of GRAB~\cite{GRAB:2020} and real images of HO3D~\cite{hampali2020honnotate}. We construct training pairs of input image, history poses and GT by corrupting the ground truth poses with synthetic errors mimicking real refiner behavior. We crop the input image around the object and input object depth estimated from MoGe-3~\cite{kong2026finedetailmonoculargeometryestimation}. At inference time, $\mathbf{h}_t$ comes from the refiner's own outputs on the last $H$ frames it refined. We recover from tracking failures, e.g., by re-registering the object. See more details in \cref{sec:supp_tracking}. The resulting trajectory stays fixed during human refinement.

\subsection{Hierarchical Interaction Enhancement}
\label{sec:method_refinement}
Human estimation and object tracking alone do not produce a coherent 4D interaction trajectory. We address this with two learned refiners in a hierarchical way: \reachnet{} first adjusts the coarse body, arms, and wrist poses against the fixed object and \touchnet{} then improves the wrist orientation and hand articulation to form correct hand-object contacts. Both reason from local geometric relations and contact evidence from images.

\subsubsection{\reachnet{}: body and wrist refinement}

The initial reconstruction can capture the overall interaction while still placing the human away from the object or penetrating it. We therefore introduce \reachnet{}, \textbf{Re}finement \textbf{a}round \textbf{c}ontact \textbf{h}ints, which leverages 3D proximity, image-based contact evidence, and temporal context to reason about the human-object relation at the coarse level of body, arm, and wrists. This contact-aware refinement encourages the reconstructed human to remain close to the object whenever supported by the observation.

\noindent\textbf{Geometry sensing with contact hints.} From the image, initialized human and object poses, we construct geometry tokens for the body and arms by defining anchors on the human and finding its closest points on the object surface. Let $\mathbf{F}_b$ and $\mathbf{F}_a$ denote the body and arm tokens, respectively. The body anchors comprise the pelvis and body joints, surface samples distributed over the torso, legs, and head, and patches along the arms. Each arm group contains anchors at the shoulder, elbow, and wrist, the finger joints and fingertips, a hand-summary anchor, and samples on the hand surface. For each human anchor $\mathbf{p}_i$, we query its closest point $\mathbf{q}_i$ on the reconstructed object and the corresponding object-surface normal $\mathbf{n}_i$. The local human--object geometry is encoded via Fourier feature encoding $\gamma(\cdot)$~\cite{li2021learnable}:
\begin{equation}
    \mathbf{g}_i = \bigl[\gamma(\mathbf{p}_i),\,\gamma(\mathbf{q}_i-\mathbf{p}_i),\,\gamma(\mathbf{n}_i)\bigr],
\end{equation}
This representation exposes both the spatial location of an anchor and its displacement and orientation relative to the object surface. Geometry alone, however, cannot determine whether two nearby surfaces should be in contact. We hence use the HOI-DETR contact flag $c_t^s$ defined in \cref{sec:method_init}. We encode this flag using a learned contact embedding $\mathbf{e}_{\mathrm{con}}$, initialized to zero, and add $c_t^s\mathbf{e}_{\mathrm{con}}$ to every token of hand $s$. Body tokens do not receive this embedding. Collecting the body geometry tokens yields $\mathbf{F}_b$, while augmenting the arm geometry tokens with this image-based contact evidence yields $\mathbf{F}_a$.

\noindent\textbf{\reachnet{} architecture.} We first map $\mathbf{F}_b$ and $\mathbf{F}_a$ to a common feature dimension with an MLP and then concatenate them into a feature tensor of shape $T\times (N_b+N_a)\times D$, where $N_b, N_a$ denotes the number of body and arm tokens respectively and $D$ is the feature dimension. We pass this through several layers of spatiotemporal attention blocks where each block performs self attention first on temporal dimension $T$ and then on spatial dimension $N_b+N_a$. These features are then sent to two prediction heads to output delta updates on the body poses $\pose^b$ and arm poses $\pose^a$.

\noindent\textbf{Training \reachnet{}.}
We build training pairs by running SAM~3D~Body~\cite{yang2026sam} and HOI-DETR~\cite{darkhalil2026improving} on GRAB~\cite{GRAB:2020} and BEHAVE~\cite{bhatnagar22behave}, and transferring the predicted local body pose onto the ground-truth body so that the global shape and translation remain the same as ground truth. The object is perturbed by small smooth random noise.
Each of the three refinement iterations is trained with:
\begin{equation}
\mathcal{L}=\mathcal{L}_{\mathrm{rot}}+\lambda_v\mathcal{L}_{\mathrm{vert}}+\lambda_t\mathcal{L}_{\mathrm{temp}}+\lambda_r\mathcal{L}_{\mathrm{ray}}+\lambda_{c}\mathcal{L}_{\mathrm{con}},
\label{eq:loss_reach}
\end{equation}
where $\mathcal{L}_{\mathrm{rot}}$ and $\mathcal{L}_{\mathrm{vert}}$ are the L2 losses on the rotations and mesh vertices of refined body and arm against ground truth and $\mathcal{L}_{\mathrm{temp}}$ is the L1 loss of their velocity and acceleration against ground truth. $\mathcal{L}_{\mathrm{ray}}$ penalizes motion across the camera ray of each input anchor, which keeps the image projection while leaving depth free. $\mathcal{L}_{\mathrm{con}}$ pulls fingertips onto the object where the ground-truth hand touches it. We remove the loss terms for arm when the dataset does not provide accurate hand annotations such as BEHAVE. Loss weights and further details are in \cref{sec:supp_reach}.

\subsubsection{\touchnet{}: finger-level contact refinement}
\label{sec:method_gears}

After \reachnet{} has aligned the body and wrist with the object, we refine the remaining local interaction using \touchnet{}: \textbf{T}emporal \textbf{O}bject s\textbf{U}rface \textbf{C}ontact for \textbf{H}ands. \touchnet{} takes the coarse hand estimate, the fixed object surface, and the per-hand HOI-DETR contact flag $c^s_t$ as input and predicts corrections to the wrist orientation and articulated fingers, while keeping the wrist position fixed.

\noindent\textbf{Local surface tokens.}
For each hand, we define $J_h$ anchors comprising its finger joints and fingertips. For every anchor, we query its $M=64$ nearest points on the object surface. Each point is represented by a seven-dimensional feature: its 3D position relative to the anchor, its 3D surface normal, and a binary indicator of whether its distance to the anchor is smaller than 4\,cm. The relative position and normal are expressed in the anchor's local joint frame. The resulting per-frame local surface features for the left and right hands are $\mathbf{F}_L,\mathbf{F}_R\in\mathbb{R}^{J_h\times M\times 7}$.

\noindent\textbf{\touchnet{} model.}
We use a PointNet encoder to aggregate the $M$ object-surface samples of each anchor in $\mathbf{F}_L$ or $\mathbf{F}_R$ into a local surface feature, which is concatenated with an embedding of the anchor position in the wrist frame. We further add a learnable embedding of the binary hand contact flag from HOI-DETR and hand visible flag from SAM~3D~Body to this feature. 
Two spatio-temporal transformer blocks then alternate self-attention over time, along each anchor across the window, with self-attention over the anchors of each frame, so every prediction uses context from both past and future frames. Prediction heads regress a position offset for every finger joint and fingertip and, from the mean token of each frame, an axis-angle update of the wrist orientation. We add the offsets to the input joints and use inverse kinematics to convert the refined joint configuration into SMPL-X hand-pose parameters $\pose^h$. This residual formulation preserves the coarse hand estimate while correcting its local alignment with the object.

\noindent\textbf{\touchnet{} training.} We train \touchnet{} with one hand per 90-frame window. The input wrist is \reachnet{}'s output placed at the ground-truth wrist position, and the object stays at its ground-truth pose. We add random rotation noise to the fingers and wrist and simulate hand occlusions with random masking. At inference time, hidden frames receive fingers interpolated between the nearest visible frames. 
We train the model with L2 loss on positions and accelerations and surface loss based on contacts and penetrations: 
\begin{equation}
\mathcal{L}_{\mathrm{\touchnet{}}}=
\mathcal{L}_{\mathrm{pos}}+\lambda_{\mathrm{acc}}\mathcal{L}_{\mathrm{acc}}
+\lambda_\mathrm{surf}\mathcal{L}_{\mathrm{surf}},
\label{eq:loss_gears}
\end{equation}
where $\mathcal{L}_{\mathrm{pos}}, \mathcal{L}_\text{acc}$ are the L2 losses on the finger joints and fingertips in the wrist frame, and their accelerations. The surface loss penalizes penetration for all frames and pushes hand to object when contact is identified. We define contacts $y_{t\kappa}$ for six hand parts $\kappa$ as one when their GT distance to object is smaller than $\sigma$ and zero otherwise. The loss $\mathcal{L}_\text{surf}$ is calculated using the aggregated object signed distance function $d(\cdot)$ for vertices per hand part $\kappa$: $\mathcal{L}_{\mathrm{surf}}=\tfrac{1}{|\mathcal{K}|}\sum_{(t,\kappa)\in\mathcal{K}}\mathrm{Huber}\big(y_{t\kappa}d(\kappa)+(1-y_{t\kappa})\min(d(\kappa),0)\big)$, where $\mathrm{Huber}(\cdot)$ is the Huber loss. Details on signed distance aggregation and $\sigma$ are provided in \cref{sec:supp_touch}.

\begin{figure*}[!t]
    \centering
    \includegraphics[width=1.0\linewidth]{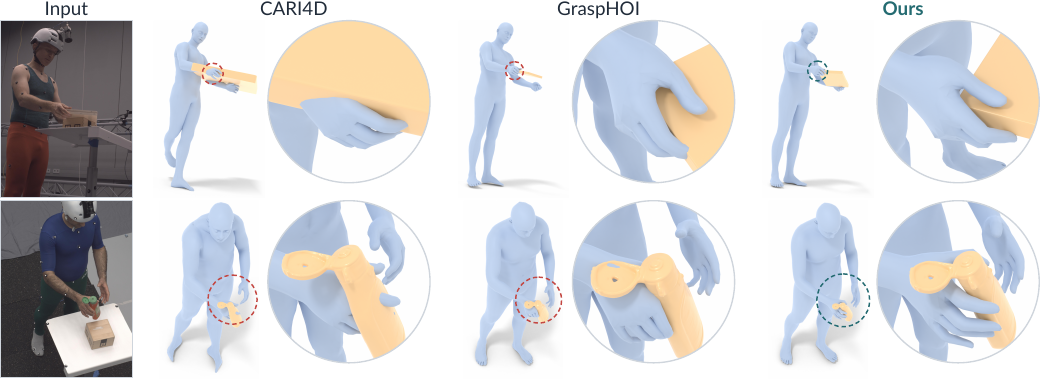}
    \caption{\textbf{Generalization to the ARCTIC dataset~\cite{fan2023arctic}, unseen in training.} CARI4D~\cite{xie2026cari4d} does not model hand articulation and GraspHOI struggles to optimize wrist pose, leading to penetrations. Our method models interaction temporally and produces accurate full body interaction together with plausible hand contacts.
    }
    \label{fig:qualitative_arctic}
\end{figure*}

\begin{figure*}[!t]
    \centering
    \includegraphics[width=1.0\linewidth]{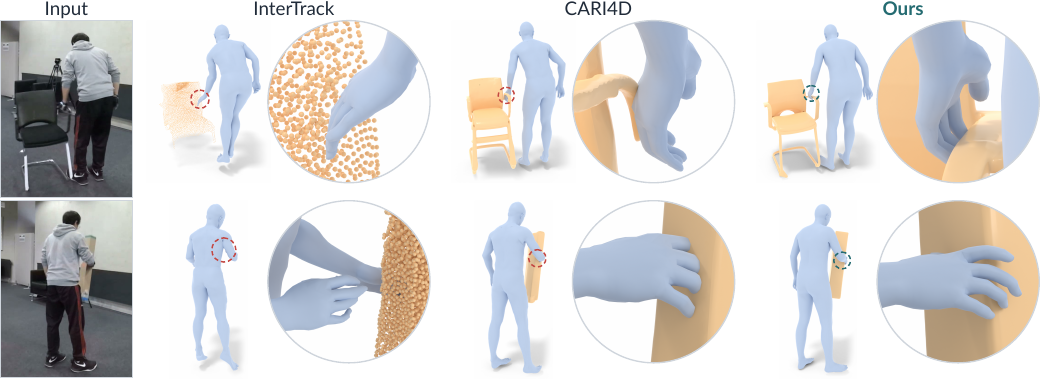}
    \caption{\textbf{Qualitative evaluation for BEHAVE dataset~\cite{bhatnagar22behave}.} InterTrack~\cite{xie2025intertrack} reconstructs object as a point cloud with hand penetration or missing contacts. Our method recovers accurate full body and object poses together with coherent hand-object surface interaction. }
    \label{fig:qualitative_behave}
    \vspace{-1mm}
\end{figure*}

\noindent\textbf{Hierarchical inference.}
At inference time, we first apply three \reachnet{} iterations to refine the body and arms against the object. We then apply \touchnet{} once to refine the wrist orientation and recover the articulated hands.

\section{Experiments}
\label{sec:experiments}

\subsection{Experimental setup}
\label{sec:exp_setup}

\begin{table*}[t]
\centering
\resizebox{\textwidth}{!}{%
\begin{tabular}{@{}lccccccccccc@{}}
\toprule
 & \multicolumn{3}{c}{Chamfer distance (cm) $\downarrow$} & \multicolumn{2}{c}{Acceleration (cm/frame$^2$) $\downarrow$} & \multicolumn{3}{c}{At contact (mm) $\downarrow$} & \multicolumn{3}{c}{Hand contact $\uparrow$} \\
\cmidrule(lr){2-4}\cmidrule(lr){5-6}\cmidrule(lr){7-9}\cmidrule(lr){10-12}
Method & Human & Object & Combined & \makebox[4.6em]{Human} & \makebox[4.6em]{Object} & Pen. & Floating & MJ2O & Precision & Recall & F1 \\
\midrule
InterTrack & 37.05 & 77.07 & 38.29 & 7.302 & 7.874 & -- & 91.93 & -- & 0.596 & 0.281 & 0.382 \\
CARI4D & 6.74 & 14.55 & 9.13 & 0.336 & 0.748 & 15.77 & 2.90 & 22.69 & 0.794 & 0.952 & 0.866 \\
Ours & \textbf{4.72} & \textbf{6.40} & \textbf{5.21} & \textbf{0.074} & \textbf{0.104} & \textbf{3.80} & \textbf{1.43} & \textbf{10.13} & \textbf{0.886} & \textbf{0.960} & \textbf{0.921} \\
\midrule
GraspHOI$^*$ & 4.85 & 11.29 & 6.84 & 5.498$^\dagger$ & 16.609$^\dagger$ & 11.97 & 4.99 & 18.76 & 0.851 & 0.948 & 0.897 \\
Ours$^*$ & \textbf{4.60} & \textbf{6.17} & \textbf{5.06} & \textbf{4.903}$^\dagger$ & \textbf{8.051}$^\dagger$ & \textbf{3.71} & \textbf{1.54} & \textbf{10.17} & \textbf{0.881} & \textbf{0.958} & \textbf{0.918} \\
\bottomrule
\end{tabular}}
\caption{\textbf{Comparison on ARCTIC~\cite{fan2023arctic}.} Contact is evaluated on hand vertices, and MJ2O on hand joints and fingertips. $^*$ denotes 1\,fps key frames only and $^\dagger$ means cm/s$^2$ for video frames at 1fps. Our method consistently outperforms baselines on all metrics. }
\label{tab:arctic_large}
\end{table*}

\begin{table*}[t]
\centering
\resizebox{\textwidth}{!}{%
\begin{tabular}{@{}lccccccccccc@{}}
\toprule
 & \multicolumn{3}{c}{Chamfer distance (cm) $\downarrow$} & \multicolumn{2}{c}{Acceleration (cm/frame$^2$) $\downarrow$} & \multicolumn{3}{c}{At contact (mm) $\downarrow$} & \multicolumn{3}{c}{Body contact $\uparrow$} \\
\cmidrule(lr){2-4}\cmidrule(lr){5-6}\cmidrule(lr){7-9}\cmidrule(lr){10-12}
Method & Human & Object & Combined & \makebox[4.6em]{Human} & \makebox[4.6em]{Object} & Pen. & Floating & MJ2O & Precision & Recall & F1 \\
\midrule
InterTrack & 12.16 & 32.67 & 18.13 & 4.510 & 3.984 & -- & 9.39 & -- & 0.856 & 0.914 & 0.884 \\
CARI4D (re-run) & 5.37 & 10.51 & 7.42 & 1.385 & 1.024 & 28.80 & \textbf{2.09} & 53.15 & 0.927 & \textbf{0.970} & \textbf{0.948} \\
Ours & \textbf{4.74} & \textbf{9.57} & \textbf{6.55} & \textbf{0.585} & \textbf{0.471} & \textbf{18.38} & 3.07 & \textbf{49.86} & \textbf{0.928} & 0.959 & 0.942 \\
\bottomrule
\end{tabular}}
\caption{\textbf{Comparison on BEHAVE~\cite{bhatnagar22behave}.} The ``At contact'' columns are measured when the ground-truth body is in contact with the object, contact scores are evaluated on whole body vertices, and MJ2O on the 22 body joints. Our method achieves more accurate human and object shapes, smoother motion, much less penetration and on par body contacts.}
\label{tab:behave_large}
\end{table*}

\noindent\textbf{Datasets.}
We train our models on GRAB~\cite{GRAB:2020} and BEHAVE~\cite{bhatnagar22behave} datasets, while evaluation is performed on the ARCTIC~\cite{fan2023arctic} dataset and the BEHAVE~\cite{bhatnagar22behave} test set. As our method assumes a rigid object, on ARCTIC we select segments of at least 150 frames in which the object's articulation, according to the ground truth, varies by at most 5$^\circ$ throughout the segment. We select two exocentric camera views for each segment, providing a total of 254 videos with 103,510 frames covering all 11 objects. On BEHAVE, we use the 62 test videos used in CARI4D~\cite{xie2026cari4d}. The history-conditioned refiner is trained on renderings of GRAB~\cite{GRAB:2020} following~\cite{xie2024template, kostyrko2026avaimg} and real images of  HO3D~\cite{hampali2020honnotate}. We train \reachnet{} on GRAB and BEHAVE and \touchnet{} on GRAB only as BEHAVE does not have accurate hands. 

\noindent\textbf{Evaluation protocol.} 
Similar to the world-space metrics in~\cite{xie2023vistracker,ye2023slahmr}, we align the reconstructed human mesh with the GT mesh within each 300-frame window and apply the same transformation to the object before computing any metrics. When necessary, we convert the predicted or GT body models to SMPL-X. We report Chamfer distance (CD) and acceleration error as defined in CARI4D~\cite{xie2026cari4d}. Following~\cite{wang2026light}, we evaluate contact using precision, recall, and F1-score. We additionally define three metrics for interaction quality: penetration, floating distance, and mean joint-to-object error (MJ2O). All three are evaluated only on frames in which the GT body is in contact with the object and are therefore grouped under ``At contact'' in the tables. Penetration and floating distance measure how much the body penetrates or floats above the object mesh, respectively. MJ2O measures the error between the predicted and GT signed distances to the object mesh at selected body joints. For ARCTIC, we use the hand joints and fingertips, whereas for BEHAVE, we use the body joints.

\noindent\textbf{Baselines.}
We compare with video based methods InterTrack~\cite{xie2025intertrack} and CARI4D~\cite{xie2026cari4d} and the image based method GraspHOI~\cite{kim2026grasphoi}. All baselines are run with the official code and pretrained models provided by the authors on shared inputs, and we leave details of baseline setups to \cref{sec:supp_exp_details}.

\subsection{Comparison with prior works}
\label{sec:exp_main}
\noindent\textbf{ARCTIC.} We report the comparison with CARI4D~\cite{xie2026cari4d}, InterTrack~\cite{xie2025intertrack} and GraspHOI~\cite{kim2026grasphoi} on ARCTIC dataset in \Cref{tab:arctic_large}. Note that this dataset is not used in training any methods hence it is an \emph{unseen} test set. We report the contact metrics on hand vertices only as this dataset contains only hand interactions. InterTrack reconstructs human as SMPL-H and object as point clouds yet its inter-frame translation is inconsistent, leading to large errors. CARI4D is designed mainly for full body interaction without fingers, and we run it with SAM~3D~Body as its human input. Nevertheless, its interaction module does not explicitly reason about fingers; hence the hand contact precision is much worse than ours. Furthermore, CARI4D struggles to reconstruct difficult viewpoints, as ARCTIC's exocentric views are often at complicated angles, such as top-down views or bottom-up views. GraspHOI~\cite{kim2026grasphoi} is an image based optimization method which is slow (47.5\,s per image); hence we compare with it only on key frames at 1fps. It reasons about hand articulation for object interaction, so the contact scores are better than CARI4D. Yet it struggles with hand occlusion and challenging viewpoints. Both GraspHOI and CARI4D try to pull the hand to the object, achieving high recall yet worse contact precision. In contrast, combined with our tracker, our coarse-to-fine refinement leads to the best result across all metrics. Notably, we reduce object Chamfer error by 56\% and improve hand contact F1 from 0.866 to 0.921 compared to CARI4D. We also show some qualitative comparisons in \Cref{fig:qualitative_arctic}.

\noindent\textbf{BEHAVE.} 
We compare against CARI4D and InterTrack on the BEHAVE~\cite{bhatnagar22behave} dataset in \Cref{tab:behave_large}. BEHAVE captures mainly full body-object interaction without articulated hand annotations; hence we report the contact metrics using body vertices and exclude comparison with GraspHOI which focuses on hand articulations. CARI4D always tries to push the hand towards object, yielding high contact recall but heavy penetrations. Our method focuses on hand contacts but achieves comparable body contact scores and is much better in acceleration and penetration errors. We also show some comparison examples in \cref{fig:qualitative_behave}. 
\begin{table*}[t]
\centering
\resizebox{\textwidth}{!}{%
\begin{tabular}{@{}lcccccccc@{}}
\toprule
 & \multicolumn{2}{c}{CD (cm) $\downarrow$} & \multicolumn{2}{c}{Accel.\ (cm/frame$^2$) $\downarrow$} & \multicolumn{3}{c}{At contact (mm) $\downarrow$} & Contact $\uparrow$ \\
\cmidrule(lr){2-3}\cmidrule(lr){4-5}\cmidrule(lr){6-8}
Method & Human & Object & \makebox[3.6em]{Human} & \makebox[3.6em]{Object} & Pen. & Floating & MJ2O & F1 \\
\midrule
a. Vanilla FoundationPose~\cite{wen2024foundationpose} & 4.81 & 18.52 & \textbf{0.067} & 0.128 & 6.21 & 47.00 & 63.52 & 0.781 \\
b. Kalman filter based FoundationPose~\cite{Wenhao_Yan_and_Jie_Chu_FoundationPose_2025} & 4.81 & 11.70 & \textbf{0.067} & 0.125 & 6.44 & 20.76 & 36.11 & 0.809 \\
c. Our history conditioned object tracking & 4.81 & 6.77 & \textbf{0.067} & 0.104 & 6.76 & 7.08 & 20.34 & 0.846 \\
\midrule
d. GRAFT~\cite{ym2026graft} (retrained) + our object pose & 10.55 & 18.26 & 1.696 & \textbf{0.103} & 5.44 & 21.26 & 43.83 & 0.671 \\
e. Our \reachnet{} w/o image contact labels & 4.75 & 6.44 & 0.070 & 0.104 & 7.67 & 4.23 & 16.03 & 0.879 \\
f. Our single-head \reachnet{} & 5.51 & 6.87 & 0.196 & 0.104 & 10.86 & 9.18 & 26.61 & 0.835 \\
g. Our full \reachnet{} & \textbf{4.72} & 6.41 & 0.074 & 0.104 & 8.25 & 1.60 & 12.16 & 0.914 \\
\midrule
h. Our \reachnet{} + \touchnet{} w/o surface loss $\mathcal{L}_\text{surf}$ & \textbf{4.72} & \textbf{6.40} & 0.074 & 0.104 & 5.79 & 1.60 & 10.57 & 0.919 \\
i. Full method & \textbf{4.72} & \textbf{6.40} & 0.074 & 0.104 & \textbf{3.80} & \textbf{1.43} & \textbf{10.13} & \textbf{0.921} \\
\bottomrule
\end{tabular}}
\caption{\textbf{Ablation studies.} We ablate our object tracking, ReachNet, and TouchNet design choices. Rows a--c pair each object tracker with the unrefined SAM~3D~Body human, and rows d--g refine the human with the same object from c.}
\label{tab:ablation}
\end{table*}

\subsection{Ablation studies}
\label{sec:exp_ablation}
We ablate our design choices on the same ARCTIC test set as described above and report the results in \cref{tab:ablation}.

\noindent\textbf{Object tracker.} We compare our history conditioned object pose tracker (\cref{tab:ablation} c) with the vanilla FoundationPose~\cite{wen2024foundationpose} (\cref{tab:ablation} a) which refines each frame independently and FoundationPose++~\cite{Wenhao_Yan_and_Jie_Chu_FoundationPose_2025} (\cref{tab:ablation} b) which uses a Kalman filter to obtain smoother and more robust tracking using FoundationPose. The human is obtained by running SAM~3D~Body~\cite{yang2026sam}. We can see that the Kalman filter stabilizes FoundationPose tracking, producing smaller object Chamfer, while our tracker, which combines learned history conditioning with failure recovery, achieves even better accuracy and smoothness.

\noindent\textbf{\reachnet{}.} We introduce \reachnet{} to reason about body and arm interaction with the object using geometric sensing and contact observation from images. We compare our \reachnet{} against GRAFT~\cite{ym2026graft} (\cref{tab:ablation} d) which refines body pose with a static scene, and also against our \reachnet{} trained without the image contact evidence (\cref{tab:ablation} e). GRAFT is retrained on the same training data as ours and all four methods use the object pose from \cref{tab:ablation} row c. We can see that single-frame based GRAFT produces large human acceleration error and missing contacts, leading to large floating error at contact and much worse human-to-object error. The image contact evidence also substantially reduces hand floating from 4.23\,mm to 1.60\,mm, showing the advantage of combining 3D geometric sensing with contact observations. We propose to use two heads to predict body and arms separately. With a single head (\cref{tab:ablation} f), the arm is less accurate, likely due to feature mixture and errors accumulated from pelvis to collar, causing shifts of the whole arm and hence worse penetration or floating.

\noindent\textbf{\touchnet{}.} \touchnet{} reasons about local object geometry and reduces the penetration and floating problems after \reachnet{} output (\cref{tab:ablation}g). Our surface loss $\mathcal{L}_\text{surf}$ is one key gradient that improves this local interaction geometry (\cref{tab:ablation}h vs. i).  
Overall, the full model (i) has the lowest human and object CD, penetration, floating, and MJ2O in the table, and the highest F1.

\begin{figure}[ht]
    \centering
    \includegraphics[width=1.0\linewidth]{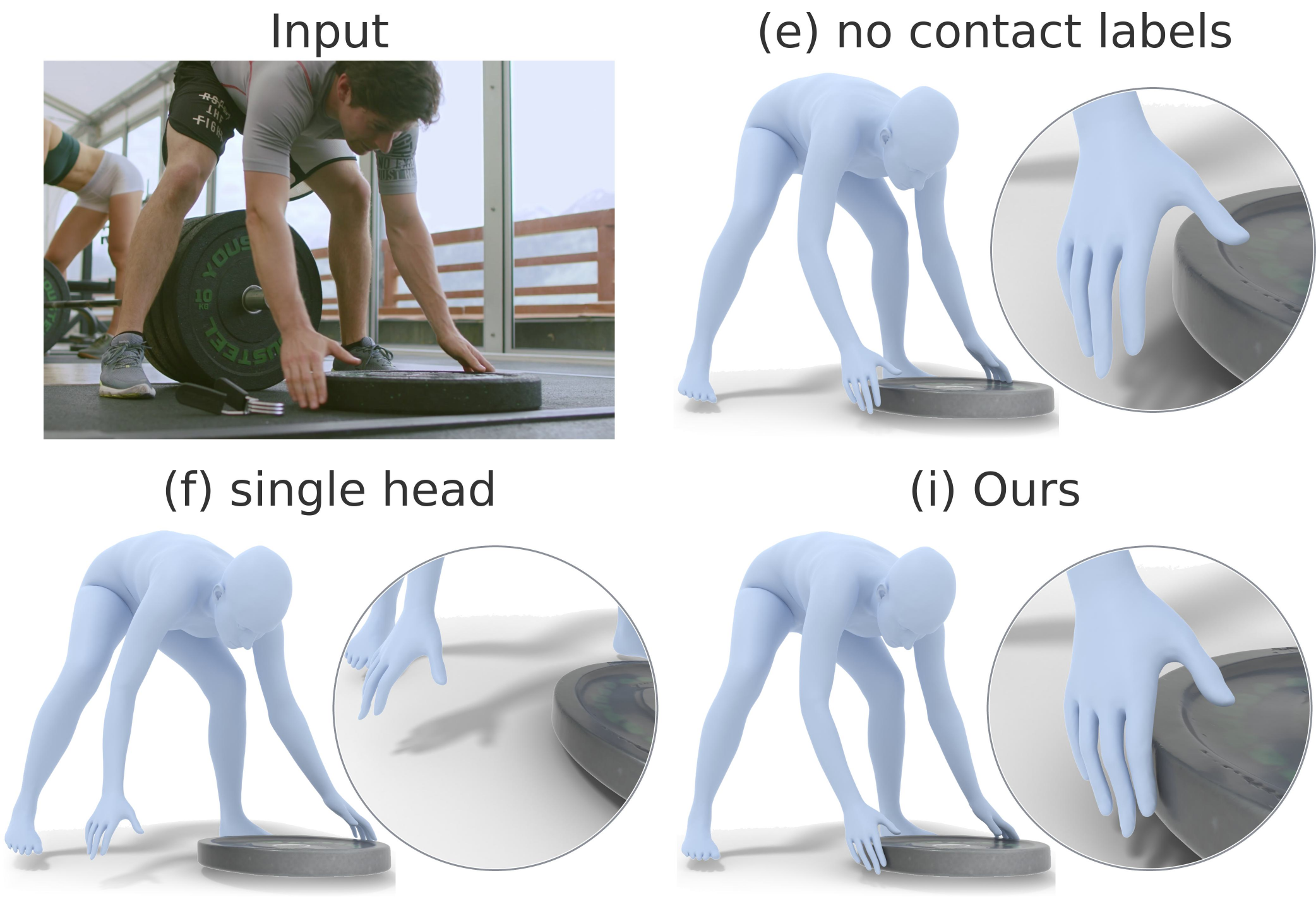}
    \caption{\textbf{Ablation examples.} Single head \reachnet{} (\cref{tab:ablation}f) predicts inaccurate wrist positions due to a swing caused by collar rotation. Without contact labels the model (\cref{tab:ablation}e) misses contact observed from the input image. }
    \label{fig:ablation}
    
\end{figure}

\noindent\textbf{Runtime comparison.} We time each method end to end on ARCTIC sequences on a single NVIDIA RTX 5090. Our method takes 1.97\,s per frame, compared with 4.83\,s for CARI4D, 12.64\,s for InterTrack and 47.5\,s for GraspHOI, making it considerably faster than all baselines. All three baselines rely on test-time optimization: GraspHOI is fully optimization-based, while InterTrack and CARI4D spend 70\% and 33\% of their runtime in optimization. In contrast, \reachnet{} and \touchnet{} are feed-forward, so most of our runtime goes to foundation models.

\section{Conclusion}
We presented \method{}, a category-agnostic method for metric 4D reconstruction of full-body human--object interactions with articulated hands from monocular RGB video. Following the natural hierarchy from whole-body approach to fine-grained grasp formation, our history-conditioned tracker stabilizes object motion under occlusion; \reachnet{} refines the body, arms, and wrists using local geometry, temporal context, and image-based contact evidence; and \touchnet{} reasons over the local object surface across time to refine wrist orientation and finger articulation. Experiments on ARCTIC and BEHAVE show improved reconstruction and interaction quality. On ARCTIC, unseen in training, \method{} reduces object Chamfer distance by 56\% and improves hand-contact F1 from 0.866 to 0.921 over CARI4D. It also applies to unseen objects and online videos. Our code and pretrained models will be released.

\paragraph{Acknowledgments.}
The project was made possible by funding from the Carl Zeiss Foundation. This work was supported by the German Federal Ministry of Education and Research (BMBF): T\"{u}bingen AI Center, FKZ: 01IS18039A. This work is funded by the Deutsche Forschungsgemeinschaft (DFG, German Research Foundation) -- 409792180 (Emmy Noether Programme, project: Real Virtual Humans). Gerard Pons-Moll is a member of the Machine Learning Cluster of Excellence, EXC number 2064/1 -- Project number 390727645. Xianghui is partially funded by the NVIDIA graduate fellowship.

{
    \small
    \bibliographystyle{ieeenat_fullname}
    \bibliography{main}
}

\clearpage

\begin{figure*}[!t]
\centering
\resizebox{\textwidth}{!}{%
\includegraphics[height=4cm]{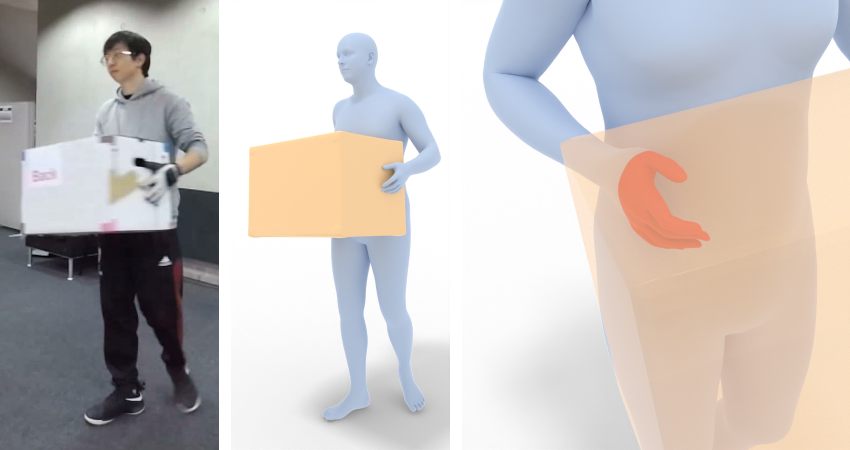}\hspace{4mm}%
\includegraphics[height=4cm]{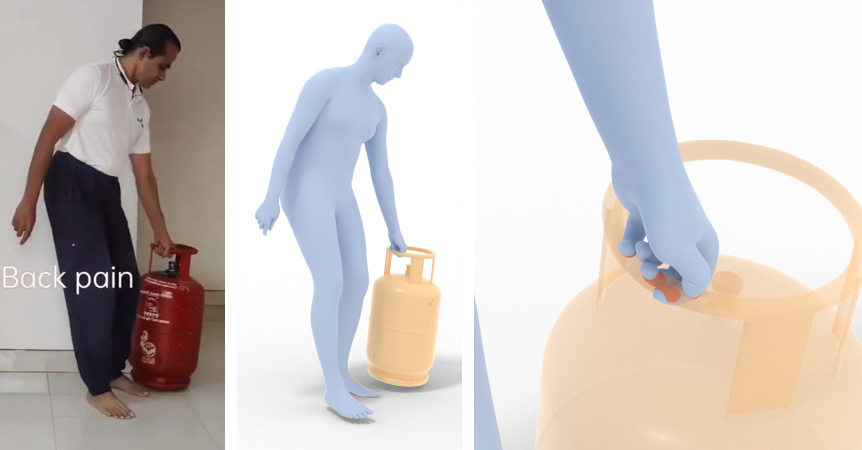}}
\caption{Failure cases, each shown as the input frame, our result from the input camera, and a close-up from another viewpoint with the object drawn semi-transparent. Left: on BEHAVE, the box hides the right hand over a long span, and the hand ends up inside the box as no information is available to infer the right hand. Right: in a video in the wild, the fingertips pass through the thin handle of a gas cylinder. Red marks the volume where the hand and the object intersect.}
\label{fig:supp_limitations}
\end{figure*}

\maketitlesupplementary
In this supplementary material, we first provide the implementation details of our method in \Cref{sec:supp_method_details}, including an additional ablation of our object tracking in \Cref{tab:fp_tracking}, and further details of our experimental setup in \Cref{sec:supp_exp_details}. We provide examples of failure cases in \Cref{fig:supp_limitations} and discussions in \Cref{sec:supp_limitations}. 
\section{Additional Method Details}\label{sec:supp_method_details}

\subsection{Initialization and Preprocessing}
\label{sec:supp_initialization}

\noindent\textbf{Contact preprocessing.}
HOI-DETR~\cite{darkhalil2026improving} detects hand and object boxes with its default score threshold of 0.3 and classifies each detected hand--object pair as interacting or not. An interacting pair is associated with our object when its object box covers at least half of $M^o_t$ (the binary object mask in frame $t$), has IoU at least 0.1 with the bounding box of $M^o_t$, and its hand box lies within a small margin of that box, proportional to the hand-box size. We assign the detection to the left or right side by projecting the estimated hand joints into the detected box. Gaps of at most five frames between two contact runs are filled as contact and remaining runs shorter than four frames are removed as non-contacting.

\noindent\textbf{Depth anchoring.}
For the depth offset $\delta_t$ of \cref{sec:method_init}, a body vertex is a ``valid sample'' if it lies within 1\,cm of the front-most rasterized body surface, projects inside $M^h_t$ and outside $M^o_t$, and has valid MoGe-3~\cite{kong2026finedetailmonoculargeometryestimation} depth. We compute the offset when at least 200 samples are visible. $\mathcal{S}_t$ holds the samples of the hands that HOI-DETR flags as contact, and those of the rest of the body when no hand is in contact. The 200-sample check is performed per hand, and if no hand is valid, the body is used. Valid offsets are smoothed into $\bar\delta_t$ by a Gaussian with a standard deviation of three frames, so that invalid frames are filled from their valid neighbors. A frame with no valid frame nearby holds the latest valid value, or the first one at the start of the video.

\noindent\textbf{Selecting the rest and contact frames.}
We identify stationary intervals by comparing each object mask with the masks one and five frames earlier and later. Object contours within a slightly dilated person mask are ignored, so that a hand entering the silhouette is not mistaken for object motion. A frame is ``stationary'' when the remaining contour moves by at most two pixels and ``moving'' when it moves by more than four pixels. Within every stationary interval, candidate $B$ maximizes the visible object area away from the person, preferring frames whose contour is mostly free of the person and the image border. Stationary intervals are ranked first by whether each interval extends at least five frames on both sides of its $B$, then by these visibility conditions, then by the availability of a contact frame (described below), and finally by the visible object area.

We extend an interval from $B$ in both directions while keeping the ``stationary'' condition, and search within this interval for contact frame $A$. This frame is sought among the stationary frames nearest to the beginning or end of object motion. Starting from the motion boundary, we select the first HOI-DETR contact whose visible hand-depth range overlaps the object-depth range. This rejects hands that merely pass in front of the object. If the chosen interval yields no contact frame, or no stationary interval exists, one frame serves as both $A$ and $B$: the largest object observation satisfying the same contact and depth-overlap test. If no frame passes this test, we use no contact frame and take the depth at $B$. $B$ remains the best rest frame, and when no stationary interval exists, becomes the frame with the largest object mask among those that HOI-DETR flags in contact, or among all frames if none is flagged. \Cref{tab:supp_init_cases} shows how often each case occurs on ARCTIC~\cite{fan2023arctic} and BEHAVE~\cite{bhatnagar22behave}.
\begin{table}[t]
\centering
\resizebox{\linewidth}{!}{%
\begin{tabular}{@{}lcccc@{}}
\toprule
 & \multicolumn{2}{c}{ARCTIC} & \multicolumn{2}{c}{BEHAVE} \\
\cmidrule(lr){2-3}\cmidrule(lr){4-5}
Case & Clips & \% & Clips & \% \\
\midrule
$A$ and $B$ found & 227 & 89.37 & 47 & 75.81 \\
No $A$ in the interval: one frame as $A$ and $B$ & 7 & 2.76 & 7 & 11.29 \\
No rest frame: one frame as $A$ and $B$ & 20 & 7.87 & 8 & 12.90 \\
No contact frame: depth at $B$ & 0 & 0.00 & 0 & 0.00 \\
\midrule
Total & 254 & & 62 & \\
\bottomrule
\end{tabular}}
\caption{How the contact frame $A$ and the rest frame $B$ are selected on the evaluation videos (\cref{sec:supp_initialization}). ``No $A$ in the interval'' means that the search in the chosen stationary interval finds no frame that passes the contact and depth-overlap test, and ``No rest frame'' that no stationary interval exists.}
\label{tab:supp_init_cases}
\end{table}

\noindent\textbf{Metric size and initial registration.}
We back-project $M^o_B$ using the anchored depth and reject points farther than three standard deviations, and additionally points whose mean distance to their 16 nearest neighbors exceeds the average by more than two standard deviations. Let $r_c$ be the maximum distance from the coordinate-wise median of the remaining point cloud and $r_m$ the corresponding radius of the canonical mesh. The initial metric scale is $s_0=r_c/r_m$, which the registration at frame $A$ below then refines.

At frame $A$, FoundationPose~\cite{wen2024foundationpose} refines 252 rotation hypotheses, comprising 42 viewpoints and six in-plane rotations, from a shared translation. FoundationPose's scoring network ScoreNet predicts the most accurate pose from the hypotheses, and RefineNet finds the delta pose and translation to fit the input image. We keep ScoreNet's top hypothesis if its rendered-mask IoU is above an IoU threshold. Otherwise, we de-duplicate hypotheses within $10^\circ$ and 0.1\,m and select the best-ranked pose whose IoU is above a second threshold, or the highest-IoU pose when none passes. We then perform 50 RGB-only refinement iterations, which allow color and silhouette evidence to correct the pose without unreliable boundary depth. This process may shift the object along the viewing ray, so we restore its depth using the median ratio between anchored depth at $A$ and rendered object depth, scaling the object around the camera center so its projection remains unchanged. This rescaling sets the object's metric scale from the anchored depth at $A$.

If the refined IoU is below an IoU threshold, we repeat scale estimation and registration at the largest object observation that passes the contact and depth-overlap test, and retain the result with higher IoU. If the final IoU is below a second threshold, we register ten sizes from $0.6s_0$ to $2s_0$ and select the one whose rendered contour best matches the visible object contour. This fallback does not apply the RGB-only refinement or the depth restoration at $A$. In our evaluation, 41 of 254 ARCTIC clips (16.1\%) and 17 of 62 BEHAVE clips (27.4\%) triggered this fallback.

\subsection{Tracking Safeguards}
\label{sec:supp_tracking}

\noindent\textbf{Kalman seeding and stillness.}
The FoundationPose++~\cite{Wenhao_Yan_and_Jie_Chu_FoundationPose_2025} filter state contains translation, Euler angles, and their velocities. At frame $t$, the filter fuses the previous pose with a 2D anchor. The anchor is the point at the object's previous depth on the camera ray through the center of the bounding box of $M^o_t$, and it measures the lateral position of the object center. The filtered pose initializes two refiner iterations, which see the depth only inside $M^o_t$. Additionally, a stationary object should not inherit depth noise. We therefore freeze its pose when the mask centroid moves by at most two pixels for at least two consecutive frames and the current mask overlaps the mask at the start of the hold with IoU at least 0.85. The filter is re-initialized on the frozen pose.

\noindent\textbf{Silhouette supervision and rescue.}
For a tracked pose ${\xi}$, we compute the occlusion-aware agreement
\begin{equation}
\mathrm{IoU}_t({\xi})=
\frac{\big|S_t({\xi})\cap M^o_t\big|}
{\big|\big(S_t({\xi})\setminus M^h_t\big)\cup M^o_t\big|},
\label{eq:iou_sup}
\end{equation}
where $S_t({\xi})$ is the rendered silhouette. Rendered pixels on the person but outside $M^o_t$ are left out, since the object may be hidden behind the person there, while rendered pixels on $M^o_t$ always count. After three consecutive frames below 0.5, FoundationPose globally re-registers on the current frame. Its hypotheses are de-duplicated within $10^\circ$ and 0.1\,m, and the 30 with the highest IoU are kept as candidates. Candidate rotations must lie within an angular bound of the last pose that passed the IoU check, a bound that grows with the number of frames between them. If candidates pass, we propose the one with the best ScoreNet score. Otherwise, we retain the current rotation and propose only the translation that maximizes IoU among the 30 candidates, because silhouettes cannot disambiguate symmetric orientations. A proposal is accepted only if it improves IoU by more than 0.05. A rescue during a stationary hold replaces the pose of the whole held interval.

\noindent\textbf{Contact-depth slide.}
A tracked object can drift along the camera ray, where monocular depth is least reliable. After each tracked frame, when all hands flagged in contact lie beyond a distance threshold from the object surface, measured from their SAM~3D~Body~\cite{yang2026sam} hand joints, we slide the object along the camera ray through its center, keeping its rotation and 2D projected position. Among log-spaced scalings of its camera distance, we take the one that brings the nearest flagged hand closest, apply it only if that hand then lies within the threshold, and keep it only if the IoU of \cref{eq:iou_sup} does not drop. The refiner and the filter then continue from the moved pose.

\noindent\textbf{Occlusion fill.}
With a clip-wide sweep, poses with occlusion-aware IoU below 0.3 are replaced by interpolating rotation spherically and camera-frame translation linearly between the nearest kept poses, holding the nearest kept pose at the ends of the video. On this trajectory, we also flag a pose when the center of the tracked object lies more than an arm's length behind the pelvis in depth, or when about 20\% or less of the object pixels rendered at the previous tracked pose fall on $M^o_t$. Flagged intervals separated by at most two frames are merged, and each interval must contain at least two consecutive flagged frames. Rotation is interpolated spherically between the trusted neighbors. For translation, we interpolate the object-center offset relative to the pelvis, which prevents the object from drifting behind the person while fully occluded. Boundary gaps hold the rotation and human-relative offset of their nearest trusted neighbor.

\noindent\textbf{History features.}
In the motion-history feature of \cref{sec:method_track}, the object translation $\boldsymbol{\tau}_s$ is the camera-frame position of the center of the object's bounding box. We use the normalizers of FoundationPose, $\rho=20^\circ$, and $r$ is half the diameter of the metric object mesh. Each history step also carries its validity flag, so the history branch reads $\nu_{t-k}[\boldsymbol{\delta}_{t-k},1]$ for $k=1,\dots,H$. As the history keeps the last $H=90$ refined poses, the oldest step $\boldsymbol{\delta}_{t-H}$, which also needs the pose at $t-H-1$, is never valid, so at most 89 steps are valid, and missing steps are zero.

\noindent\textbf{History-head training.} We train on 7,200 frames from renderings of GRAB~\cite{GRAB:2020} training clips and 3,895 real frames from HO3D~\cite{hampali2020honnotate} sequences, with MoGe-3 depth shifted in each frame by its median difference from the ground-truth object depth over the visible object pixels. Each sample is drawn from GRAB with probability 0.6 and from HO3D with probability 0.4. As at test time, the refiner sees FoundationPose's crop around the starting pose, with depth only inside the visible object mask. With probabilities 0.35, 0.30, and 0.35, the starting pose is: (1) the ground-truth pose of a frame 1 to 30 frames earlier, jittered by up to $3^\circ$ and 5\,mm; (2) the ground-truth pose moved along its camera ray by up to $2r$ and rotated by up to $10^\circ$; or (3) the ground-truth pose under a FoundationPose-style random perturbation of up to $15^\circ$, $0.15r$ laterally, and $0.5r$ in depth, 5\% of which are large rotations of $20^\circ$ to $180^\circ$. Starting poses more than $90^\circ$ from the ground truth are left out of the loss. To teach the refiner when to trust depth, a black box slides over the object with probability 0.3, removing the depth under it, and with probability 0.25 the input depth over the object is shifted away from the camera by $0.5r$ to $8r$. The history holds the ground-truth poses of the preceding 90 frames, corrupted by a slowly drifting error simulated with an AR(1) process with coefficient 0.98 and stationary standard deviations of $3.6^\circ$, capped at $15^\circ$, and $0.02r$ per axis. It is dropped with probability 0.15, with two exceptions: it is always kept when the black box is applied, and for 80\% of the samples with starting pose (2), it is replaced by a valid history without motion. The loss is the L1 distance between the predicted and ground-truth pose updates in the normalized units of FoundationPose, with the depth translation weighted twice the lateral ones. We use AdamW with batch size 16, learning rate $10^{-5}$ for the released refiner and $10^{-4}$ for the history branch, weight decay $10^{-4}$, an L2-SP penalty of $10^{-3}$ toward the released weights, and gradient clipping at norm 5. The batch-normalization running statistics keep their released values, and we use the checkpoint after 20,000 samples.

\begin{table}[t]
\centering
\small
\setlength{\tabcolsep}{4pt}
\begin{tabular}{@{}lcc@{}}
\toprule
 & \multicolumn{2}{c}{Object} \\
\cmidrule(lr){2-3}
Method & CD (cm) $\downarrow$ & Accel.\ (cm/frame$^2$) $\downarrow$ \\
\midrule
a. FoundationPose~\cite{wen2024foundationpose} & 18.52 & 0.128 \\
b. a + fine-tuned, no history & 14.92 & 0.122 \\
c. FoundationPose++~\cite{Wenhao_Yan_and_Jie_Chu_FoundationPose_2025} & 11.70 & 0.125 \\
d. c + fine-tuned, no history & 12.99 & 0.118 \\
\midrule
e. Ours w/o rescue, fill, slide & 8.96 & \textbf{0.104} \\
f. Ours & \textbf{6.77} & \textbf{0.104} \\
\bottomrule
\end{tabular}
\caption{Additional ablation on our object tracking designs on ARCTIC. ``fine-tuned, no history'' uses the refiner fine-tuned on our training data without the history branch. ``w/o rescue, fill, slide'' turns off the silhouette rescue and the contact-depth slide during tracking and the occlusion fill after it (\cref{sec:supp_tracking}).}
\label{tab:fp_tracking}
\end{table}

\subsection{Network Architectures}
\label{sec:supp_arch}
\begin{figure*}[t]
\centering
\resizebox{\textwidth}{!}{%
\begin{tikzpicture}[
  font=\footnotesize, >={Stealth[length=2mm]},
  box/.style={draw, rounded corners=2pt, align=center, minimum height=10mm, inner sep=3pt},
  blk/.style={box, fill=gray!15},
  inp/.style={box, minimum width=27mm},
  lab/.style={font=\small\bfseries, inner sep=0pt},
  dim/.style={font=\scriptsize, align=center, inner sep=1pt},
  node distance=4mm and 14mm]
\node[lab, anchor=north west] (rl) at (0,0) {ReachNet};
\node[inp, anchor=north west] (r1) at (0,-5mm) {Body tokens $\mathbf{F}_b$: 62\\Arm tokens $\mathbf{F}_a$: $2\times56$};
\node[box, right=36mm of r1] (r2) {Token MLP\\+ $c^s_t\mathbf{e}_{\mathrm{con}}$ on $\mathbf{F}_a$\\+ motion embedding};
\node[dim, above=6mm of r2] (rf) {contact flag $c^s_t$};
\node[blk, right=of r2] (r3) {$3\times$ block\\temporal attention ($\pm45$)\\spatial attention};
\node[box, right=of r3] (r4) {Body and arm heads\\with GRU decoders};
\node[dim, right=5mm of r4, align=left] (r5) {update of $\pose^b$: $(b,t,16,6)$\\update of $\pose^a$: $(b,t,2,3,6)$};
\draw[->] (r1) -- node[dim, above] {point tokens: $(b,t,166,29)$} node[dim, below] {patch tokens: $(b,t,8,8,9)$} (r2);
\draw[->] (rf) -- node[dim, right] {$(b,t,2)$} (r2);
\draw[->] (r2) -- node[dim, above] {$(b,t,$\\$174,256)$} (r3);
\draw[->] (r3) -- node[dim, above] {$(b,t,$\\$174,256)$} (r4);
\draw[->] (r4) -- (r5);
\draw[->] (r5.south) -- ++(0,-6mm) -| node[pos=0.25, fill=white, inner sep=1pt] {$\times3$ refinement iterations} (r1.south);
\node[lab, anchor=north west] (tl) at (0,-30mm) {TouchNet};
\node[inp, anchor=north west] (t1a) at (0,-35mm) {Local surface features\\$\mathbf{F}_L$ or $\mathbf{F}_R$};
\node[inp, below=5mm of t1a] (t1b) {Anchor positions\\in wrist frame};
\node[box, right=of t1a, minimum width=16mm] (t2a) {PointNet};
\node[box, right=of t1b, minimum width=16mm] (t2b) {MLP};
\node[box, anchor=west, minimum height=22mm] (t3) at ($(t2a.east)!0.5!(t2b.east)+(24mm,0)$) {Concatenate\\+ contact and visible-flag\\embeddings\\+ position encoding};
\node[dim, above=6mm of t3] (tf) {contact flag $c^s_t$,\\hand visible flag};
\node[blk, right=of t3] (t4) {$2\times$ block\\temporal attention\\spatial attention};
\node[box, right=of t4] (t5) {Prediction\\heads};
\node[dim, right=5mm of t5, align=left] (t6) {joint offsets: $(b,90,20,3)$\\wrist: $(b,90,3)$};
\draw[->] (t1a) -- node[dim, above] {$(b,90,20,$\\$64,7)$} (t2a);
\draw[->] (t1b) -- node[dim, above] {$(b,90,$\\$20,3)$} (t2b);
\draw[->] (t2a.east) -- node[dim, above] {$(b,90,20,128)$} (t2a.east -| t3.west);
\draw[->] (t2b.east) -- node[dim, above] {$(b,90,20,128)$} (t2b.east -| t3.west);
\draw[->] (tf) -- node[dim, right] {$(b,90,2)$} (t3);
\draw[->] (t3) -- node[dim, above] {$(b,90,$\\$20,256)$} (t4);
\draw[->] (t4) -- node[dim, above] {$(b,90,$\\$20,256)$} (t5);
\draw[->] (t5) -- (t6);
\end{tikzpicture}}
\caption{Architectures of ReachNet and TouchNet, with tensor shapes, where $b$ is the batch size and $t$ the number of frames. Each block applies temporal self-attention, spatial self-attention over the tokens of a frame, and a feed-forward layer.
}
\label{fig:supp_arch}
\end{figure*}
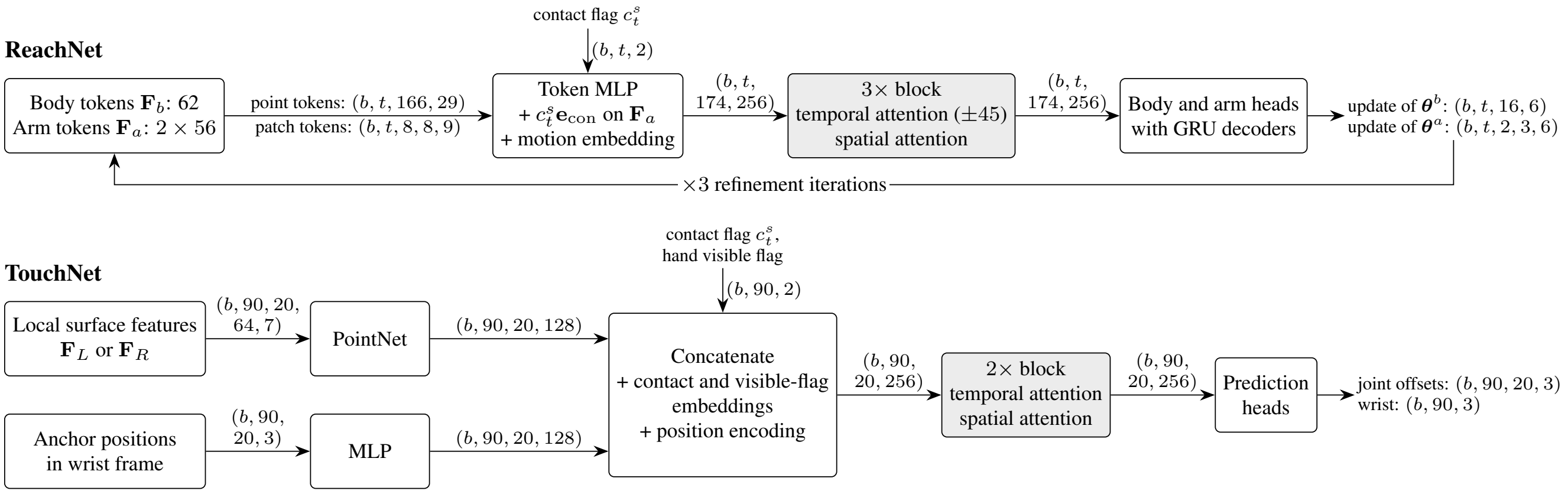

\Cref{fig:supp_arch} summarizes both networks.

\noindent\textbf{ReachNet.}
A frame holds 174 tokens. For each arm, $\mathbf{F}_a$ holds 56 tokens: the 21 hand joints, the shoulder, the elbow, a summary token, and 32 hand-surface samples. $\mathbf{F}_b$ holds 62 body tokens: 22 body joints, 32 body-surface samples, and 8 arm-surface patches. The geometry features of each token are mapped by an MLP to $D=256$ and receive a learned embedding of the token type, and every token of hand $s$ also receives the contact embedding $c^s_t\mathbf{e}_{\mathrm{con}}$. Motion features of the input over offsets of 1 to 32 frames are embedded and added to every token. Three blocks, each with eight attention heads and a feed-forward width of 1024, apply temporal self-attention within a band of $\pm45$ frames with rotary position encoding, followed by spatial self-attention over all 174 tokens of the frame. The body head predicts rotation updates for the root and the remaining body joints from the body tokens, and the arm head predicts, for each side, rotation updates for the shoulder, elbow, and wrist from the arm tokens of that side. The body and arm heads each end in a temporal decoder, a bidirectional GRU with two layers and hidden size 128, which reads the predicted updates of the whole sequence together with the motion features and adds a zero-initialized residual to them, so that the updates stay consistent over time.

\noindent\textbf{TouchNet.}
A window holds 90 frames of one hand, with 20 tokens per frame for the 15 finger joints and 5 fingertips. For each token, a shared PointNet ($7\!\to\!64\!\to\!128\!\to\!128$) encodes each of the $M=64$ object samples nearest to the joint and aggregates them by averaging the encodings over the samples within 4\,cm; a linear layer then gives a 128-dimensional surface feature, which is zero when no sample lies within 4\,cm. An MLP ($3\!\to\!64\!\to\!128$) embeds the joint position in the wrist frame, and the two features are concatenated into a 256-dimensional token. Learned embeddings of the HOI-DETR contact flag $c^s_t$ and the hand visible flag are added to every token, followed by sinusoidal encodings of the frame and joint indices. Two blocks, each with four attention heads and a feed-forward width of 1024, alternate temporal self-attention over the 90 frames with spatial self-attention over the 20 tokens of a frame. The prediction heads, a linear layer on each token and a zero-initialized MLP ($256\!\to\!64\!\to\!3$) on the mean token of each frame, output the displacement of each joint in its joint frame and an axis-angle update of the wrist orientation, capped at $30^\circ$.

\subsection{ReachNet Training}
\label{sec:supp_reach}

ReachNet is trained on GRAB~\cite{GRAB:2020} and on the training sequences of BEHAVE~\cite{bhatnagar22behave}, where BEHAVE supervises only the body terms because its hand articulation is unavailable. The input body is obtained by running SAM~3D~Body on the GRAB renderings and the BEHAVE training videos and converting it to SMPL-X. We transplant the root and body rotations of this estimate onto the ground-truth body, which keeps the ground-truth shape and pelvis translation, so the input preserves realistic estimator errors without introducing a shape mismatch. On GRAB, the input keeps the ground-truth finger articulation, which ReachNet does not change, as we want ReachNet to predict accurate wrist rotation. The HOI-DETR flags are computed on the same renderings and videos. The input object is the ground-truth object surface perturbed by temporally correlated noise, a rotation about a random axis and a translation along each axis, with stationary standard deviations of approximately $7^\circ$ and 15\,mm. The perturbed object trajectory is then smoothed over time by a Gaussian with a standard deviation of three frames. The perturbation magnitude is randomized per crop by a factor between 0.3 and 1.5, and 30\% of crops additionally drop and refill object points. Arm and contact labels follow this perturbed object, while keeping the hand rigid, so that the hand contact stays faithful to the ground truth. This process is applied so that the network learns the interaction relative to the geometry it actually receives.

The body objective combines temporal consistency, root and local-rotation supervision, body-surface supervision before and after centering, and an image-ray anchor. The arm objective supervises hand surfaces, arm rotations, elbows, and temporal consistency, together with an image-ray anchor and the contact term. \Cref{tab:supp_reach_weights} assigns each term to a group of \cref{eq:loss_reach} and lists its weight. Rotations, in the 6D representation, and positions are compared by squared distances, while $\mathcal{L}_{\mathrm{temp}}$ compares first and second temporal differences by an L1 distance. On BEHAVE, the body-surface terms leave out the arms. For a point ${a}$ displaced by ${d}$ from its input position and its unit input camera ray ${\ell}$, the image anchor is
\begin{equation}
\mathcal{L}_{\mathrm{ray}}(\mathcal{X})=
100\operatorname*{mean}_{t,{a}\in\mathcal{X}}
\left\lVert{d}-({d}^{\top}{\ell}){\ell}\right\rVert^2,
\end{equation}
which preserves image alignment while allowing correction along the ambiguous depth direction. For the body, $\mathcal{X}$ holds the body joints except the pelvis, elbows, and wrists, and 32 body-surface samples. For each arm, $\mathcal{X}$ holds the elbow and the 21 hand joints, weighted by the visibility of the hand. The contact loss attracts a fingertip region only when its labeled position lies within 3\,mm of the input object, with frames detected by HOI-DETR weighted ten times more strongly. It penalizes only floating.

\begin{table}[t]
\centering
\small
\begin{tabular}{@{}llcc@{}}
\toprule
Group & Term & Weight & Data \\
\midrule
$\mathcal{L}_{\mathrm{rot}}$ & root orientation & 5 & both \\
 & body joint rotations & 2 & both \\
 & arm rotations & 5 & GRAB \\
\midrule
$\mathcal{L}_{\mathrm{vert}}$ & body vertices & 7 & both \\
 & centered body vertices & 5 & both \\
 & hand vertices & 7 & GRAB \\
 & centered hand vertices & 5 & GRAB \\
 & elbow positions & 3 & GRAB \\
\midrule
$\mathcal{L}_{\mathrm{temp}}$ & body joints and rotations & 1 & both \\
 & arm and hand & 1 & GRAB \\
\midrule
$\mathcal{L}_{\mathrm{ray}}$ & body anchors & 1 & both \\
 & arm and hand anchors & 1 & GRAB \\
\midrule
$\mathcal{L}_{\mathrm{con}}$ & fingertip contact & 2 & GRAB \\
\bottomrule
\end{tabular}
\caption{Weights of the ReachNet loss terms, listed by their group in \cref{eq:loss_reach}, applied in every refinement iteration. Terms marked ``both'' are used on GRAB and BEHAVE, and terms marked ``GRAB'' on GRAB only, as these losses are not applicable to BEHAVE due to its limited annotation. The ``body/hand vertices'' terms place the body and hand, and the ``centered'' terms keep the shape and orientation errors from being outweighed when the hand is in the wrong place, where a large position error would otherwise dominate the plain vertex error.}
\label{tab:supp_reach_weights}
\end{table}

ReachNet is trained from scratch for 37,500 steps with Adam. The learning rate warms linearly to $10^{-4}$ over 2,000 steps and decays cosinusoidally to $2\times10^{-7}$, and gradients are clipped at norm 1, separately for the body branch and for the remaining weights. Training uses one refinement iteration for the first 2,500 steps, two until step 5,000, and three afterwards, and the losses of all iterations are summed. Gradients are stopped between refinement iterations, so each iteration starts from the detached result of the previous one. Within an iteration, one pass of the shared network feeds both heads. The arm terms supervise the update of the arm head at the shoulder position before the body update, and re-attaching the arm to the moved shoulder is computed without gradient, while the contact term sees the root and body rotations detached. The arm, hand, and contact terms therefore train the shared network and the arm head, but not the body head. In each crop, the geometry features $\mathbf{g}_i$ of the shoulder and elbow anchors of each arm are zeroed with probability 0.35, so that the network relies less on the estimated arm position. Training crops contain at least 90 frames and at most 256 frames, and each step draws $\lfloor 360/T\rfloor$ crops of $T$ frames, with $T=90$ with probability 0.75 and $T$ uniform in $[91,256]$ otherwise.

\subsection{TouchNet Training and Post-processing}
\label{sec:supp_touch}

TouchNet is trained on 90-frame windows from GRAB. Its coarse fingers are the SAM~3D~Body estimate positioned at the ground-truth wrist, with the wrist rotation predicted by ReachNet. Visible finger rotations are moved along the geodesic from the SAM~3D~Body estimate through the ground truth, occasionally overshooting the ground truth by up to 50\% of that movement, so that excessive penetration or floating can be simulated. They also receive a random rotation of $12^\circ$ RMS. We simulate hidden hands with probability 0.3 using a span whose length is geometrically distributed with a mean of nine frames and, with probability 0.15, an additional span of 30--90 frames. Frames that the hand visible flag marks as hidden on the renderings are hidden as well, and hidden frames receive interpolated fingers as at test time.

For the surface term in \cref{eq:loss_gears}, the signed distance of each output hand vertex is its signed tangent-plane distance $s=n_p^\top(v-p)$ to the nearest valid local-surface sample $p$ of the anchors at the ends of its bone. Vertices more than 5\,mm from the normal line through $p$ are ignored. For each frame and each of six hand parts---the palm and five fingers---let $d_{t\kappa}$ be the smallest signed distance (the $d(\kappa)$ of \cref{sec:method_gears} in frame $t$) and $\hat d_{t\kappa}$ its ground-truth counterpart, so that $y_{t\kappa}=1$ when $\hat d_{t\kappa}\le\sigma=3$\,mm. We use
\begin{equation}
\mathcal{L}_{\mathrm{surf}}=
\operatorname*{mean}_{(t,\kappa)\in\mathcal{K}}
\begin{cases}
\mathrm{Huber}(d_{t\kappa}), & \hat d_{t\kappa}\le3\,\mathrm{mm},\\
\mathrm{Huber}\big(\min(d_{t\kappa},0)\big), & \text{otherwise},
\end{cases}
\end{equation}
with a 1\,mm Huber knee, where $\mathcal{K}$ holds the part-frames whose part has at least one vertex that is not ignored. Thus a contacting part is pulled to the surface whether it floats or penetrates, while a non-contacting part is only pushed out. The surface loss is ramped in over 5,000 steps, and we use $\lambda_{\mathrm{acc}}=0.9$ and $\lambda_{\mathrm{surf}}=9.8$, so that the acceleration term takes 10\% and the surface loss 40\% of the initial loss. We train for 40,000 steps with Adam, batch size 8, learning rate $5\times10^{-4}$ with cosine annealing, and gradient clipping at norm 1.

Finally, we remove isolated finger-rotation spikes. A frame is flagged as spiked when a joint rotates sharply away from both temporal neighbors up to three frames away, while those neighbors remain much closer to one another. All three joints of the affected finger are replaced over the spike by spherical interpolation, and detection and replacement run for up to three passes. If TouchNet lifts an already-contacting hand so that its nearest vertex floats more than 3\,mm from the surface, we interpolate its wrist and finger rotations back toward the ReachNet pose until its nearest vertex lies just inside that distance. These rules use no ground truth and do not move the wrist position or object, and the last one acts on each hand and frame independently. On ARCTIC, the spike removal changes 127 of 103{,}510 frames (0.12\%).

\section{Additional Experimental Details}\label{sec:supp_exp_details}

\noindent\textbf{Shared inputs and scoring.}
All methods read the same person and object masks. On BEHAVE, these are the masks released with the dataset, as in prior work~\cite{xie2026cari4d, xie2025intertrack}. On ARCTIC, all methods likewise receive the same object and person masks. All outputs are scored with the protocol of \cref{sec:exp_setup}. SMPL-H bodies, from the baselines and from the BEHAVE ground truth, are converted to SMPL-X with the deformation transfer released with SMPL-X~\cite{pavlakos2019expressive}, followed by a per-frame fit. We use the Chamfer distance and acceleration error definitions of CARI4D~\cite{xie2026cari4d} and the evaluation protocol of VisTracker~\cite{xie2023vistracker}. Unlike CARI4D, which fits a single transformation to the first frame of each video, VisTracker fits it to all frames of each window, so that one unreliable frame cannot misalign an entire video.

\noindent\textbf{CARI4D.}
We run the official code with its released checkpoint. As the human input, we replace NLF~\cite{sarandi2024neural} with SAM~3D~Body~\cite{yang2026sam}. On ARCTIC, we follow its pipeline for videos in the wild, while its FoundationPose tracking runs in both temporal directions as in its BEHAVE pipeline. Evaluation uses the same object meshes as ours, scaled by its own metric-scale estimation. On BEHAVE, we re-run its official BEHAVE pipeline, again with SAM~3D~Body as the human input, using the object meshes, calibrated intrinsics, and 2D keypoints released by the CARI4D authors.

\noindent\textbf{InterTrack.}
We use the official code and released checkpoints without retraining. On ARCTIC, we run the authors' InterCap script with translation estimated from 2D, after warping every frame and mask into the BEHAVE Kinect camera that InterTrack assumes, using the calibrated intrinsics of each view. We use its general reconstruction model, as ARCTIC's objects are outside its chair-specific one, and the five-object pose network of that script. InterTrack reconstructs each frame up to scale, so before the window alignment we divide each frame by InterTrack's estimated body scale. On BEHAVE, for Chamfer distance and acceleration, each frame is instead rescaled by the predicted human scale, as in the authors' evaluation, before the same window alignment. As its object is a point cloud, the object Chamfer distance uses its points, penetration and MJ2O are undefined, and floating and contact use the distance from the hand or body vertices to the nearest object point.

\noindent\textbf{GraspHOI.}
GraspHOI~\cite{kim2026grasphoi} optimizes single images, so we run it on the 1\,fps samples of ARCTIC, every 30th frame of each video, skipping the dark first frame in 118 videos, for 3,382 frames in total. It reads our inputs: SAM~3D~Body, the HOI-DETR detections, the masks, MoGe-3 depth and intrinsics. We score our method on the same 3,382 frames (Ours$^*$). On these samples, each 300-frame window holds at most 10 samples and its alignment is fitted on them, and accelerations are taken between samples 1\,s apart.

\section{Limitations}\label{sec:supp_limitations}
\noindent\textbf{Heavy occlusions.}
When a hand stays heavily occluded by the object over a long span, SAM~3D~Body has little evidence for it and often places it deep inside the object. ReachNet may then keep the hand close to this estimate, or move it only partway out, leaving the wrist away from its true position. TouchNet does not fix this either, as it refines the wrist orientation and the fingers around the wrist that ReachNet places and cannot pull out a hand that already penetrates deeply. Such hands can therefore remain inside the object over the whole occluded span. This happens only under very long and heavy occlusions as our temporal module can reason short occlusions from nearby frames. Recovering these hands under long occlusions would require more generative capability of the model, which we leave to future work.

\noindent\textbf{Thin objects.}
TouchNet refines the fingers from the positions and normals of object surface samples near each joint. A thin object or part, such as a blade, a handle, or a sheet, offers only a few such samples around each finger, and their normals are less reliable, as the two sides of the part lie close together. With so little local geometry, TouchNet cannot reliably tell where the surface is or which side of it a finger is on, so fingers may float or pass through the part. One major reason is the training data, as most objects in GRAB, a common choice for learning hand--object interaction, are compact. Future work could learn such local grasps from more recent hand--object datasets with more thin objects, or from grasps simulated for robotic manipulation.

\end{document}